\documentclass[11pt,letterpaper]{article}
\usepackage{emnlp2017}
\usepackage{times}
\usepackage{latexsym}
\usepackage[colorinlistoftodos]{todonotes}
\usepackage{amsmath}
\usepackage{url}
\usepackage{graphicx}
\usepackage{caption}
\usepackage{subcaption}
\usepackage{kantlipsum}
\usepackage{comment}
\usepackage{hyperref}
\usepackage{makecell}
\usepackage{multirow}

\emnlpfinalcopy

\def\emnlppaperid{1223}

\title{Video Highlight Prediction Using Audience Chat Reactions}

\author{Cheng-Yang Fu, Joon Lee, Mohit Bansal, Alexander C. Berg \\
UNC Chapel Hill\\
  {\tt \{cyfu, joonlee, mbansal, aberg\}@cs.unc.edu}}

\date{}

\begin{document}

\maketitle

\begin{abstract}

Sports channel video portals offer an exciting domain for research on multimodal, multilingual analysis.  
We present methods addressing the problem of automatic video highlight prediction based on joint visual features and textual analysis of the real-world audience discourse with complex slang, in both English and traditional Chinese. We present a novel dataset based on League of Legends championships recorded from North American and Taiwanese Twitch.tv channels (will be released for further research), and demonstrate strong results on these using multimodal, character-level CNN-RNN model architectures.
\end{abstract}

\section{Introduction}
On-line eSports events provide a new setting for observing large-scale social interaction focused on a visual story that evolves over time---a video game. While watching sporting competitions has been a major source of entertainment for millennia, and is a significant part of today's culture, eSports brings this to a new level on several fronts.  One is the global reach, the same games are played around the world and across cultures by speakers of several languages.  Another is the scale of on-line text-based discourse during matches that is public and amendable to analysis. One of the most popular games, League of Legends, drew 43 million views for the 2016 world series final matches  (broadcast in 18 languages) and a peak concurrent viewership of 14.7 million\footnote{\scriptsize{\url{http://www.lolesports.com/en_US/articles/2016-league-legends-world-championship-numbers}}}. Finally, players interact through what they see on screen while fans (and researchers) can see {\em exactly} the same views.

This paper builds on the wealth of interaction around eSports to develop predictive models for match video highlights based on the audience's online chat discourse as well as the visual recordings of matches themselves.  ESports journalists and fans create highlight videos of important moments in matches.  Using these as ground truth, we explore automatic prediction of highlights via multimodal CNN+RNN models for multiple languages.  Appealingly this task is {\em natural}, as the community already produces the ground
 truth and is global, allowing multilingual multimodal grounding.

\begin{figure}
    \centering
	\includegraphics[width=0.45\textwidth]{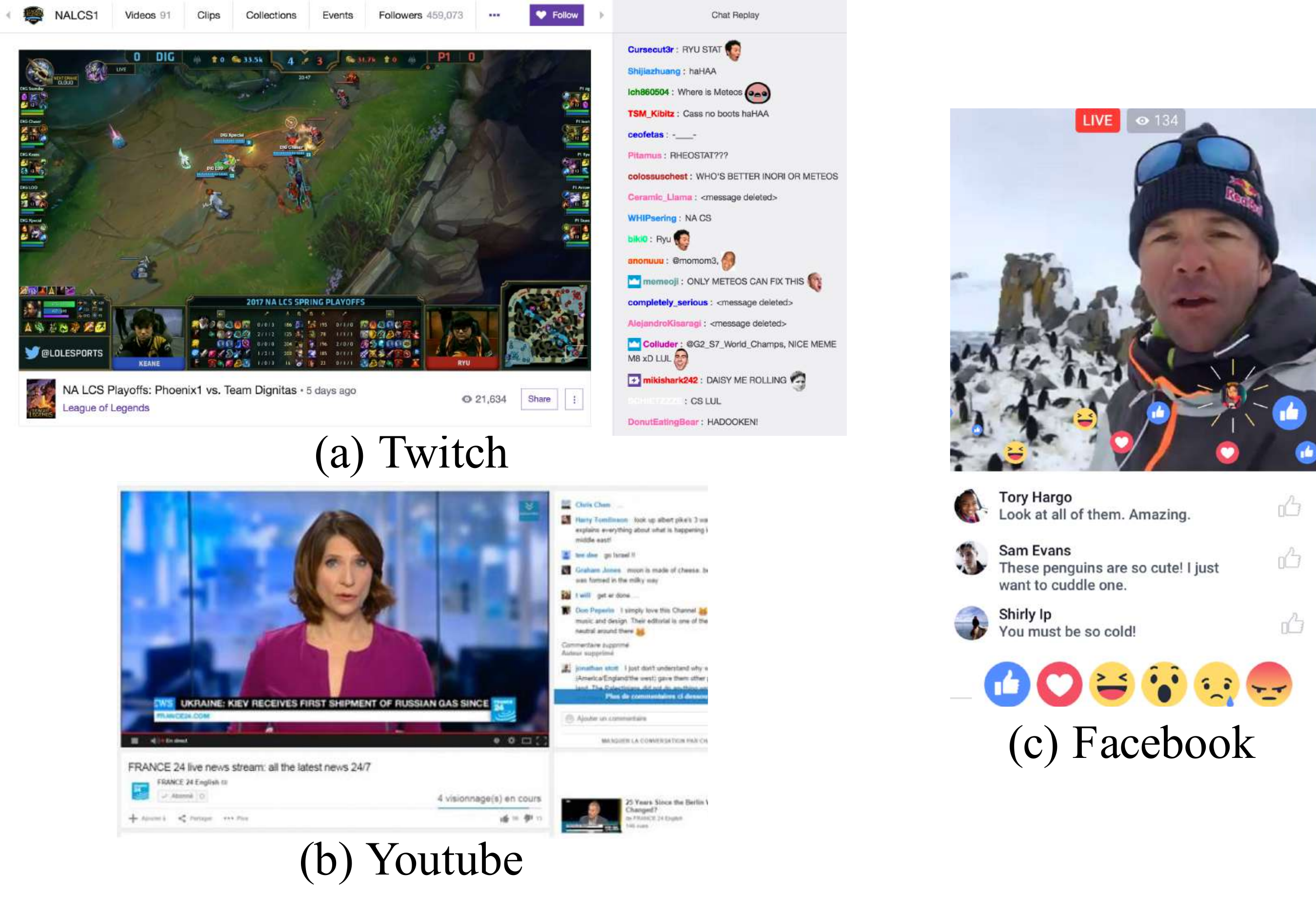}
    \caption{Pictures of Broadcasting platforms:(a) Twitch: League of Legends Tournament Broadcasting, (b) Youtube: News Channel, (c)Facebook: Personal live sharing}
    \label{fig:broadcasting}
\end{figure}

Highlight prediction is about capturing the exciting moments in a specific video (a game match in this case), and depends on the context, the state of play, and the players.  This task of predicting the exciting moments is hence different from summarizing the entire match into a story summary.  Hence, highlight prediction can benefit from the available real-time text commentary from fans, which is valuable in exposing more abstract background context, that may not be accessible with computer vision techniques that can easily identify some aspects of the state of play. As an example, computer vision may not understand why Michael Jordan's dunk is a highlight over that of another player, but concurrent fan commentary might reveal this.

We collect our dataset from Twitch.tv, one of the live-streaming platforms that integrates comments (see Fig.~\ref{fig:broadcasting}), and the largest live-streaming platform for video games. We record matches of the game League of Legends (LOL), one of the largest eSports game in two subsets, 1) the spring season of the North American League of Legends Championship Series (NALCS), and 2) the League of Legends Master Series (LMS) hosted in Taiwan/Macau/HongKong, with chat comments in English and traditional Chinese respectively.  We use the community created highlights to label each frame of a match as highlight or not.

In addition to our new dataset, we present several experiments with multilingual character-based models, deep-learning based vision models either per-frame or tied together with a video-sequence LSTM-RNN, and combinations of language and vision models.  Our results indicate that while {\em surprisingly} the visual models generally outperform language-based models, we can still build reasonably useful language models that help disambiguate difficult cases for vision models, and that combining the two sources is the most effective model (across multiple languages).

\section{Related Work}
\label{sec:related_works}
We briefly discuss a small sample of the related work on language and vision datasets, summarization, and highlight prediction.  There has been a surge of vision and language datasets focusing on captions over the last few years, \cite{pascal_caption,im2text,MSCOCO}, followed by efforts to focus on more specific parts of images \cite{visualgenome}, or referring expressions \cite{ReferItGame}, or on the broader context \cite{visualstory}.  For video, similar efforts have collected descriptions \cite{chen:acl11}, while others use existing descriptive video service (DVS) sources~\cite{MovieDescription,MovieDescription2}. Beyond descriptions, other datasets use questions to relate images and language \cite{{VQA},VisualMadlibs}.  This approach is extended to movies in~\newcite{MovieQA}.  

The related problem of visually summarizing videos (as opposed to finding the highlights) has produced datasets of holiday and sports events with multiple users making summary videos~\cite{SumMe} and multiple users selecting summary key-frames \cite{VSUMM} from short videos.  For language-based summarization, \textit{Extractive models} \cite{NLP_summary_est1,NLP_summary_est2} generate summaries by selecting important sentences and then assembling these, while \textit{Abstractive models} \cite{NLP_summary_abs,NLP_summary_selective,nallapati2016abstractive,see2017get} generate/rewrite the summaries from scratch. 

Closer to our setting, there has been work on highlight prediction in football (soccer) and basketball based on audio of broadcasts \cite{highlight_audio_basketball} \cite{highlight_audio_soccer} where commentators may have an outsized impact or visual features \cite{highlight_soccer}.  In the spirit of our study, there has been work looking at tweets during sporting events \cite{highlight_twitter}, but the tweets are not as immediate or as well aligned with the games as the eSports comments.  More closely related to our work,~\newcite{yahoo_esports} collects videos for Heroes of the Storm, League of Legends, and Dota2 on online broadcasting websites of around 327 hours total. They also provide highlight labeling annotated by four annotators. Our method, on the other hand, has a similar scale of data, but we use existing highlights, and we also employ textual {\em audience} chat commentary, thus providing a new resource and task for Language and Vision research.  
In summary, we present the first language-vision dataset for video highlighting that contains audience reactions in chat format, in multiple languages. The community produced ground truth provides labels for each frame and can be used for supervised learning. The language side of this new dataset presents interesting challenges related to real-world Internet-style slang.

\section{Data Collection}
\label{sec:data_collection}
 Our dataset covers 218 videos from NALCS and 103 from LMS for a total of 321 videos from week 1 to week 9 in 2017 spring series from each tournament. Each week there are 10 matches for NALCS and 6 matches for LMS. Matches are best of 3, so consist of two games or three games. The first and third games are used for training. The second games in the first 4 weeks are used as validation and the remainder of second games are used as test. Table~\ref{table:data split} lists the numbers of videos in train, validation, and test subsets. 
 
\begin{table}[hb!]
	\centering
    \setlength{\tabcolsep}{6pt}
    \begin{tabular}{l| c | c |c|c } 
    
    Dataset &  Train &  Val  & Testing & Total\\
    \hline
  	NALCS &   128  & 40 & 50 & 218\\
    \hline
    LMS & 57 & 18 & 28 & 103 \\
	\end{tabular}
    \caption{Dataset statistics (number of videos).}
    \label{table:data split}
\end{table}

Each game's video ranges from 30 to 50 minutes in length which contains image and chat data linked to the specific timestamp of the game. The average number of chats per video is 7490 with a standard deviation of 4922. The high value of standard deviation is mostly due to the fact that NALCS simultaneously broadcasts matches in two different channels (nalcs1\footnote{\scriptsize{\url{https://www.twitch.tv/nalcs1}}} and nalcs2\footnote{\scriptsize{\url{https://www.twitch.tv/nalcs2}}}) which often leads to the majority of users watching the channel with a relatively more popular team causing an imbalance in the number of chats. If we only consider LMS which broadcasts with a single channel, the average number of chats are 7210 with standard deviation of 2719. The number of viewers for each game averages about 21526, and the number of unique users who type in chat is on average 2185, i.e., roughly 10\% of the viewers.

\begin{figure}

        \begin{subfigure}[t]{0.4\linewidth}
            \centering
            \includegraphics[width=\linewidth]{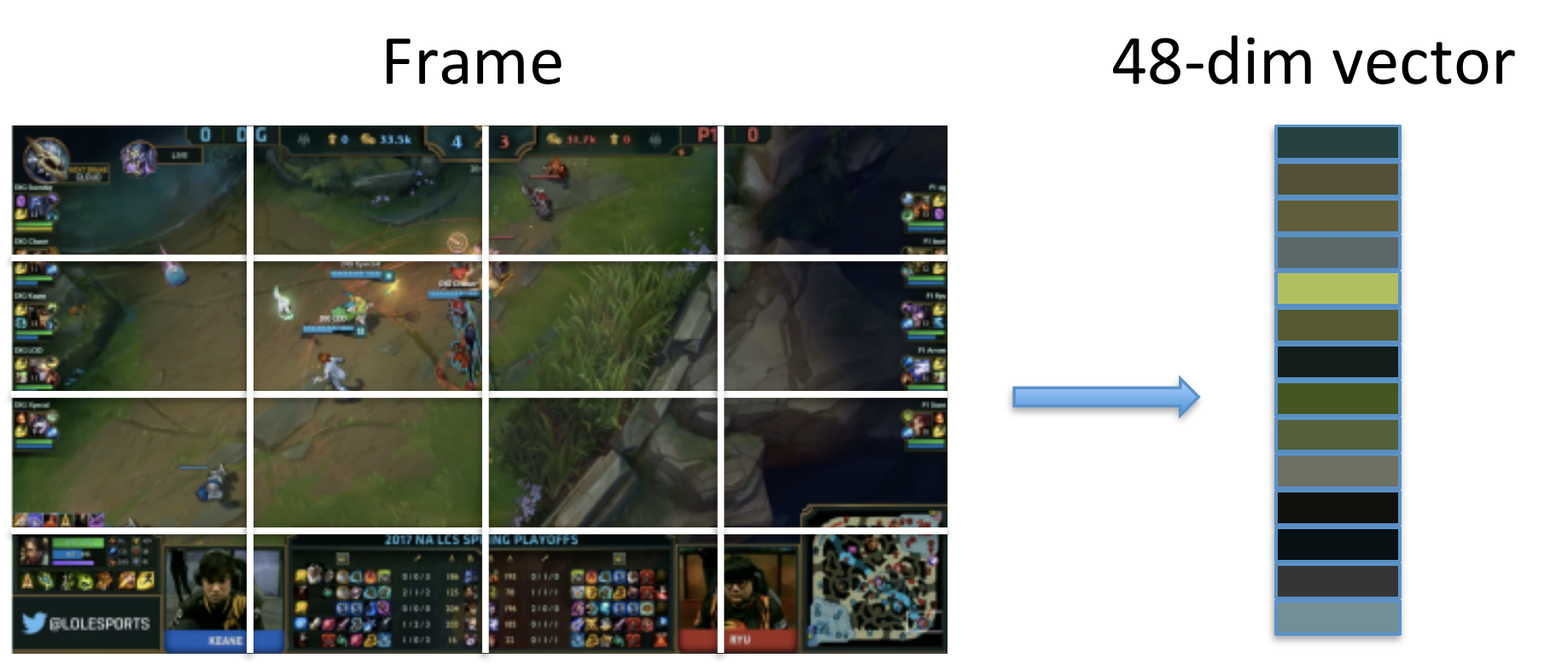}
            \caption{Feature vector of frame}
            \label{fig:MatchingFeature}
        \end{subfigure}
        \hfill
        \begin{subfigure}[t]{0.45\linewidth}
            \centering
            \includegraphics[width=\linewidth]{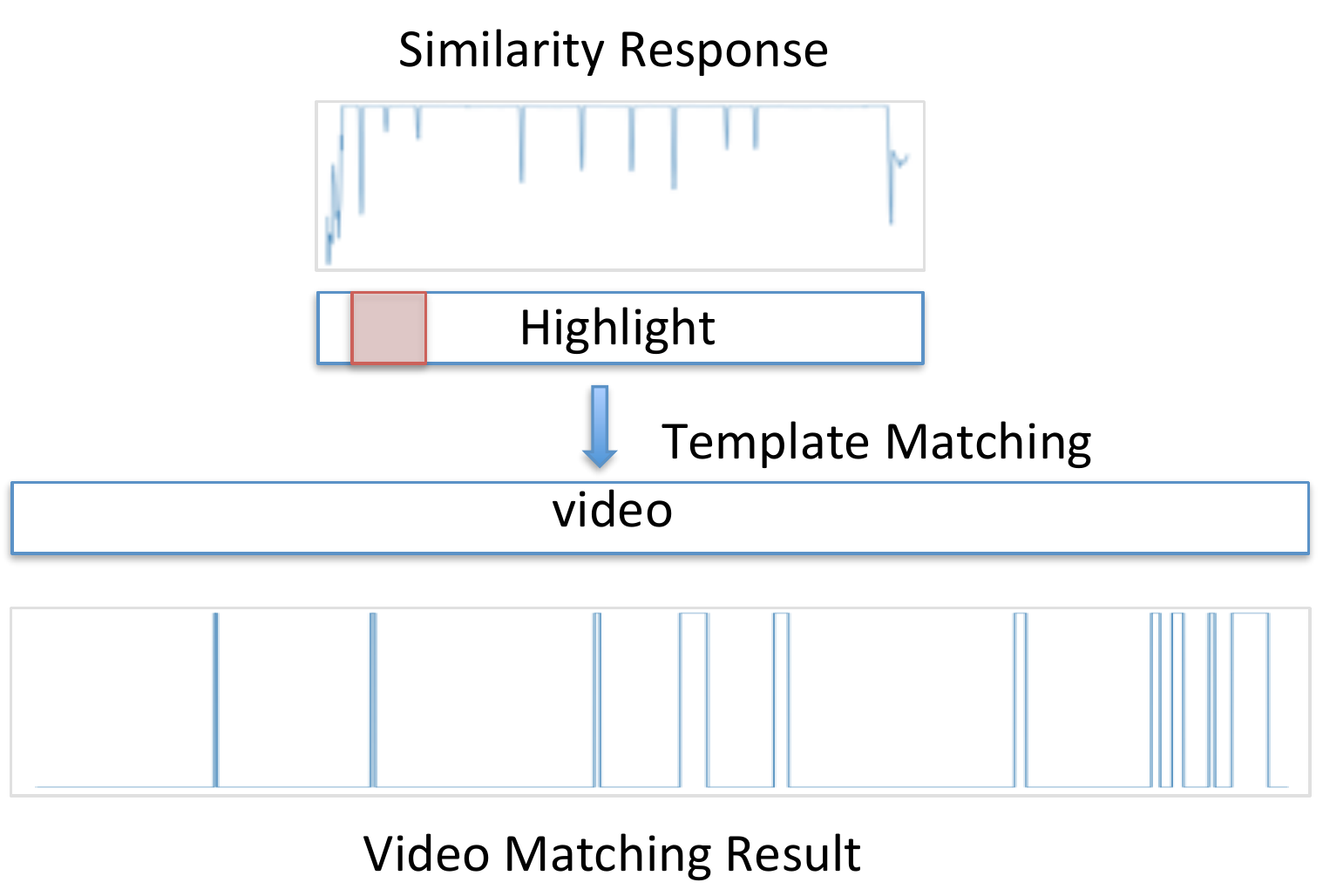}
            \caption{Template Matching}
            \label{fig:MatchingFig}
        \end{subfigure}
     \caption{Highlight Labeling: (a) The feature representation of each frame is calculated by averaging each color channel in each subregion. (b) After template matching, the top bar shows the maximum of similarity matching of each frame in the highlight and the bottom bar is the labeling result of the video.}
\end{figure}

\paragraph{Highlight Labeling} For each game, we collected community generated highlights ranging from 5 minutes to 7 minutes in length. For the purpose of consistency within our data, we collected the highlights from a single Youtube channel, Onivia,\footnote{\scriptsize{\url{https://www.youtube.com/channel/UCPhab209KEicqPJFAk9IZEA}}} which provided highlights for both championship tournaments in a consistent arrangement. We expect such consistency will aid our model to better pick up characteristics for determining highlights.
We next need to align the position of the frames from the highlight video to frames in the full game video. For this, we adopted a template matching approach. For each frame in the video and the highlight, we divide it into 16 regions of 4 by 4 and use the average value of each color channel in each region as the feature. The feature representation of each frame ends up as a 48-dim vector as shown in Figure \ref{fig:MatchingFeature}. For each frame in the highlight, we can find the most similar frame in the video by calculating distance between these two vectors. However, matching a single frame to another suffers from noise. Therefore, we alternatively concatenate the following frames to form a window and use template matching to find the best matching location in the video. We found out that when the window size is 60 frames, it gives consistent and high quality results. For each frame, the result contains not only the best matching score but also the location of that match in the video.\footnote{When the window contains a moment of clip transition in highlights, the best matching score appears low. This is used to separate all clips in the highlight. Then we can use the starting and end locations of each clip to label the video.}
Figure \ref{fig:MatchingFig} illustrates this matching process.

\begin{figure*}[tb!]
        \begin{subfigure}[b]{0.15\linewidth}
            \centering
            \includegraphics[width=\linewidth]{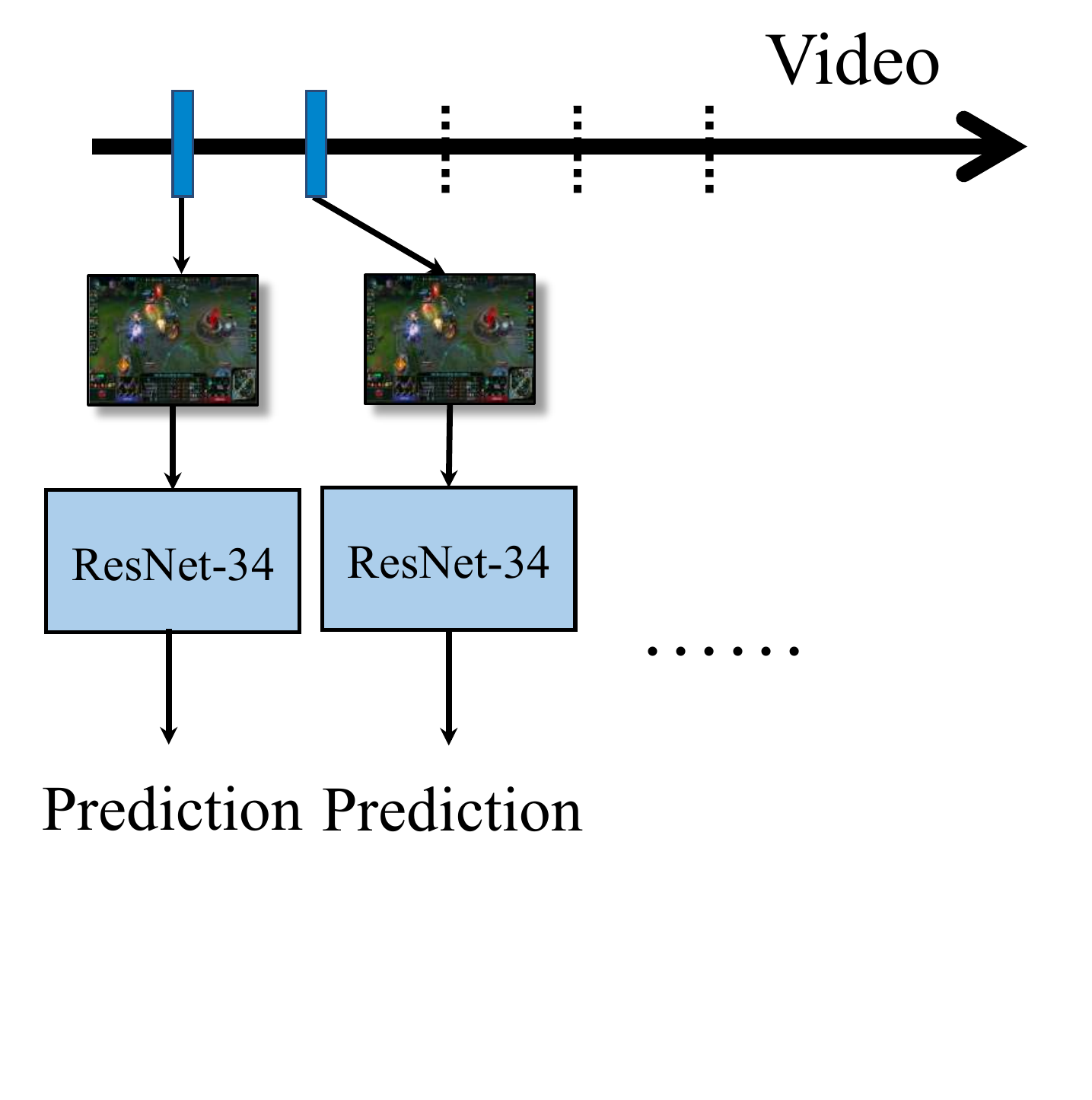}
            \caption{{V-CNN}}
            \label{fig:frameCNN}
        \end{subfigure}
        \hfill
        \begin{subfigure}[b]{0.20\linewidth}
            \centering
            \includegraphics[width=\linewidth]{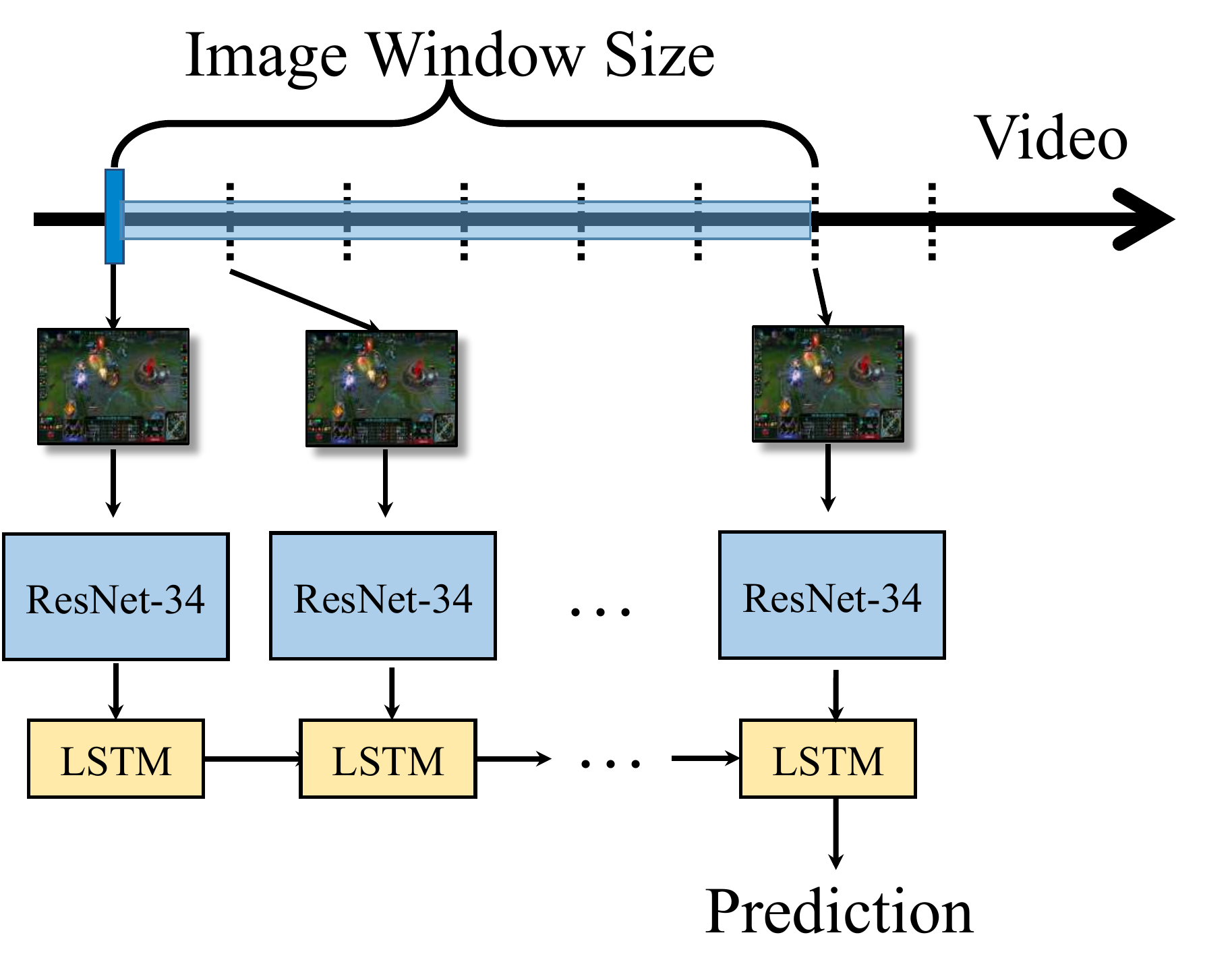}
            \caption{{V-CNN-LSTM}}
            \label{fig:LSTMCNN}
        \end{subfigure}
        \hfill
        \begin{subfigure}[b]{0.26\linewidth}
            \centering
            \includegraphics[width=\linewidth]{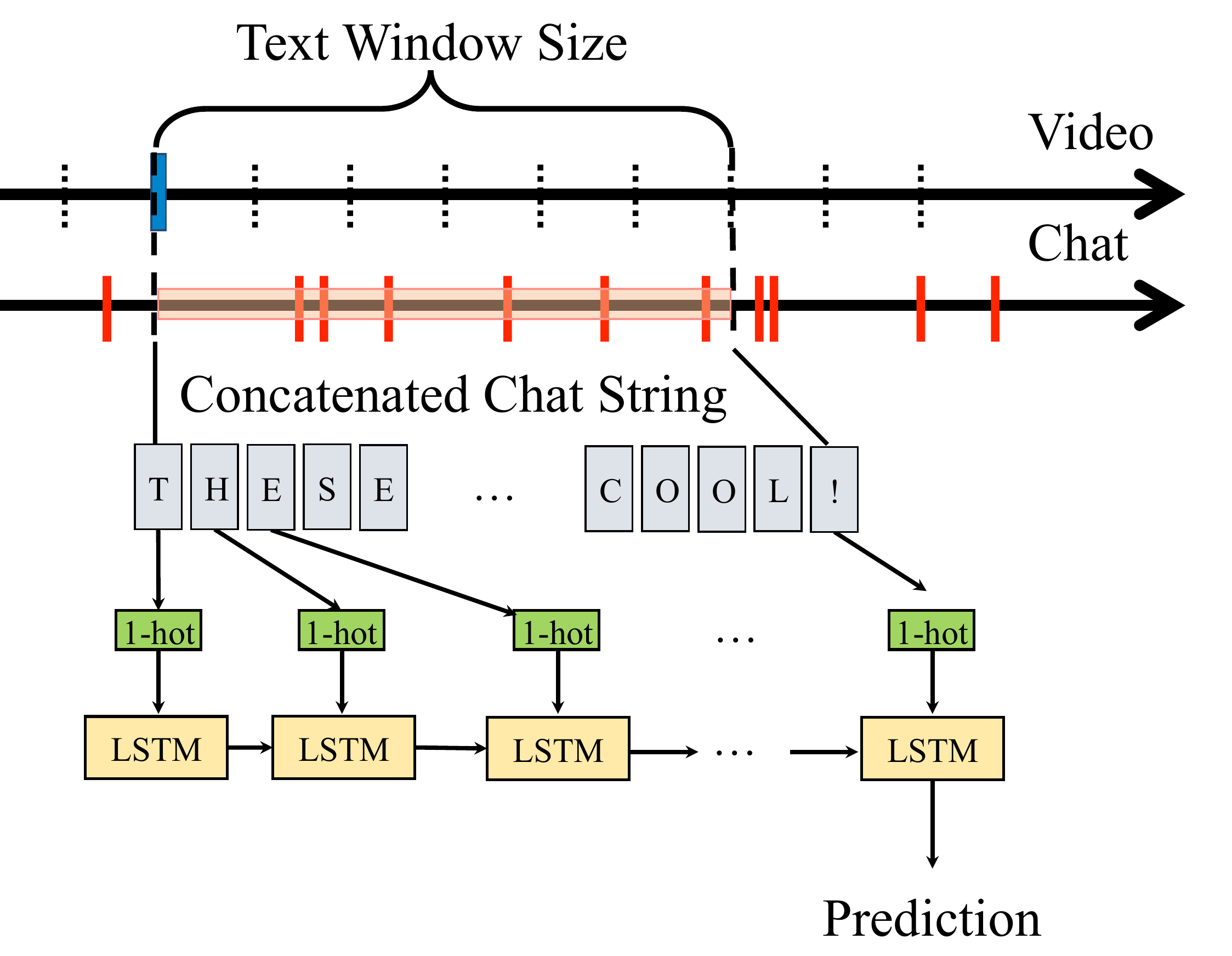}
            \caption{{L-Char-LSTM}}
            \label{fig:charLSTM}
        \end{subfigure}
        \hfill
        \begin{subfigure}[b]{0.33\textwidth}
            \centering
            \includegraphics[width=\linewidth]{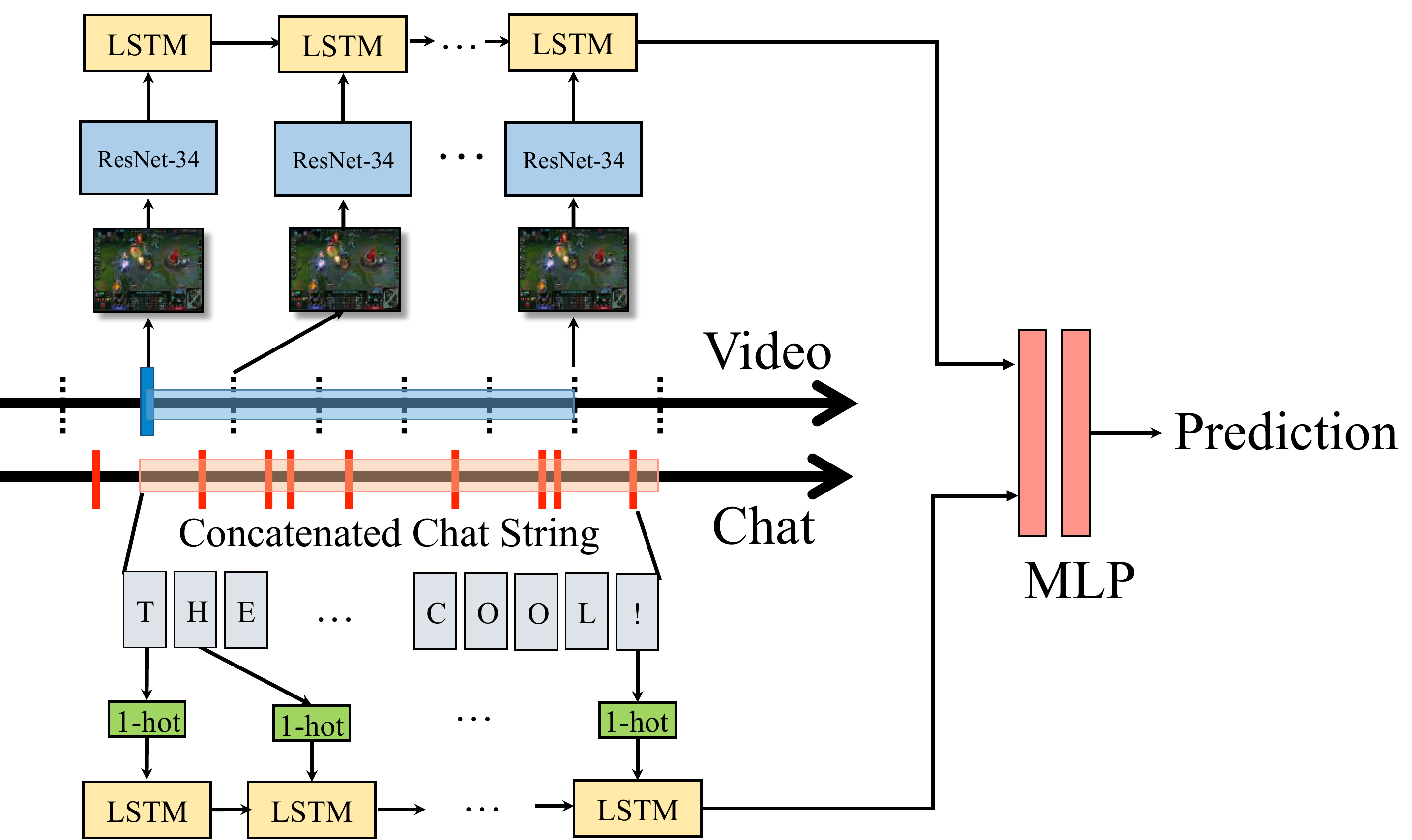}
          \caption{Full model : $lv$-LSTM}
            \label{fig:Multimodal}
        \end{subfigure}
   \vspace{-5pt}
        \caption{
        Network architecture of proposed models.} 
        \label{fig:models}
\vspace{-5pt}
\end{figure*}

\section{Model} 
\label{sec:model}
In this section, we explain the proposed models and components. We first describe the notation and definition of the problem, plus the evaluation metric used. Next, we explain our vision model V-CNN-LSTM and language model L-Char-LSTM. Finally, we describe the joint multimodal model $lv$-LSTM.

\paragraph{Problem Definition}
Our basic task is to determine if a frame of the full input video should be labeled as being part of the output highlight or not.  To simplify our notation, we use $X=\{x_1, x_2, ..., x_t\}$ to denote a sequence of features for frames.  Chats are expressed as $C=\{(c_1, ts_1), ... ,(c_n, ts_n) \}$. where each chat $c$ comes with a timestamp $ts$.  Methods take the image features and/or chats and predict labels for the frames,  $Y=\{y_1, y_2, ..., y_t\}$.\\
{\bf Evaluation Metric:} We refer to the set of frames with positive ground truth label as $S_{gt}$ and the set of predicted frames with a positive label as $S_{pred}$. Following \cite{SumMe,TVSum}, we use the harmonic mean F-score in Eq.\ref{eq:fmeasure} widely used in video summarization task for evaluation:
\begin{align}
P &= \frac{ S_{gt} \cap S_{pred}}{|S_{pred}|} , \quad  R = \frac{S_{gt} \cap S_{pred}}{ |S_{gt}|} \\
F &= \frac{2PR}{P+R} \times 100 \% 
\label {eq:fmeasure}
\end{align}

\paragraph{V-CNN}
We use the ResNet-34 model~\cite{ResidualNetwork} to represent frames,
motivated by its strong results on the ImageNet
Challenge~\cite{ILSVRC15}.  Our naive V-CNN model (Figure~\ref{fig:frameCNN}) uses features
from the pre-trained version of this
network \footnote{\scriptsize{\url{https://github.com/pytorch/pytorch}}} directly
to make prediction at each frame (which are resized to 224x224).

\paragraph{V-CNN-LSTM} 
In order to exploit visual video information sequentially over time, we use a memory-based LSTM-RNN on top of the image features, so as to model long-term dependencies. All of our videos are 30FPS. As the difference between consecutive frames is usually minor, we run prediction every 10th frame during evaluation 
and interpolate predictions between these frames. During training, due to the GPU memory constraints, we unfold the LSTM cell 16 times. Therefore the image window size is around 5-seconds (16 samples every 10th frame from 30fps video). The hidden state from the last cell is used as the V-CNN-LSTM feature. This process is shown in Figure~\ref{fig:LSTMCNN}.

\paragraph{L-Word-LSTM and L-Char-LSTM}
Next, we discuss our language-based models using the audience chat text. Word-level LSTM-RNN models~\cite{wordLSTM} are a common approach to embedding sentences. Unfortunately, this does not fit our Internet-slang style language with irregularities, ``mispelled" words (hapy, happppppy), emojis (\char94\_\char94), abbreviations (LOL), marks (?!?!?!?!), or onomatopoeic cases (e.g., “4” which sounds like “yes” in traditional Chinese). People may type variant length of “4”, e.g.,, “4444444” to express their remarks.

Therefore, alternatively, we model the audience chat with a character-level LSTM-RNN model \cite{Char-LSTM}. Characters of the language, Chinese, English, or Emojis, are expanded to multiple ASCII characters according to the two-character Unicode or other representations used on the chat servers.  We encode a 1-hot vector for each ASCII input character. For each frame we use all chats that occur in the next $W_t$ seconds which are called text window size to form the input for L-Char-LSTM. We concatenate all the chats in a window, separating them by a special stop character, and then fed to a 3-layer L-Char-LSTM model.\footnote{The number of these stop characters is then an encoding of the number of chats in the window. Therefore, the L-Char-LSTM could learn to use this \#chats information, if it is a useful feature. Also, some content has been deleted by Twitch.tv or the channel itself due to the usage of improper words. We use symbol "$\backslash$n" to replace such cases.} This model is shown in Figure~\ref{fig:charLSTM}.
Following the setting in Sec.~\ref{sec:ablation}, we evaluate the text window size from 5 seconds to 9 seconds, and got the following accuracies:32.1\%, 29.6\%, 41.5\%, 28.2\%, 34.4\%. We achieved best results with text window size as 7 seconds, and used this in rest of the experiments.

\paragraph{Joint \MakeLowercase{$Lv$}-LSTM Model}
Our final $lv$-LSTM model combines the best vision and language models: V-CNN-LSTM and L-Char-LSTM. 
For the vision and language models, we can extract features $F_{v}$ and $F_{l}$ from V-CNN-LSTN and L-Char-LSTM, respectively. Then we concatenate $F_v$ and $F_l$, and feed it into a 2-layer MLP. The completed model is shown in Figure~\ref{fig:Multimodal}. We expect there is room to improve this approach, by using more involved representations, e.g., Bilinear Pooling \cite{VQABilinear}, Memory Networks \cite{memorynetwork_VQA}, and Attention Models \cite{attention_VQA}; this is future work.

\section{Experiments and Results}
\label{sec:experiment}

\begin{table}[t]
    \setlength{\tabcolsep}{6pt}
    \begin{tabular}{l| c | c | c |c|c}
    
    Method & Data & UF & P & R &F\\
    \hline
    \scriptsize{L-Char-LSTM} & C& \footnotesize{100\%} & 0.11 & 0.99 &19.6 \\ 
    \hline
    \scriptsize{L-Char-LSTM} & C& \footnotesize{last 25\%} & 0.35& 0.51 & 41.5 \\
    \hline
    \scriptsize{L-Word-LSTM} & C& \footnotesize{last 25\%} & 0.10 & 0.99 & 19.2\\
    \hline
    \scriptsize{V-CNN} & V& \footnotesize{100\%} &  0.40 & 0.93 & 56.2 \\ 
    \hline
    \scriptsize{V-CNN} & V & \footnotesize{last 25\%} & 0.57 & 0.74 & 64.0 \\ 
    \hline
    \scriptsize{V-CNN-LSTM} & V & \footnotesize{last 25\%} & 0.58 & 0.82 & 68.3 \\ 
    \hline 
    \scriptsize{$lv$-LSTM} & C+V & \footnotesize{last 25\%} & 0.77 & 0.72 & \textbf{74.8} \\
    \hline 
	\end{tabular}
    \vspace{-5pt}
    \caption{Ablation Study: Effects of various models. \textbf{C}:Chat, \textbf{V}:Video, \textbf{UF}: \% of frames Used in highlight clips as positive training examples; \textbf{P}: Precision, \textbf{R}: Recall, \textbf{F}: F-score.}
    \vspace{-10pt}
    \label{tab:ablation}
\end{table}

\paragraph{Training Details}

In development and ablation studies, we use train and val splits of the data from NALCS to evaluate models in Section~\ref{fig:models}. For the final results, models are retrained on the combination of train and val data (following major vision benchmarks e.g. PASCAL-VOC and COCO), and performance is measured on the test set. We separate the highlight prediction to three different tasks based on using different input data: videos, chats, and videos+chats. The details of dataset split are in Section~\ref{sec:data_collection}. Our code is implemented in PyTorch. 

To deal with the large number of frames total, we sample only 5k positive and 5k negative examples in each epoch.  We use batch size of 32 and run 60 epochs in all experiments. Weight decay is $10^{-4}$ and learning rate is set as $10^{-2}$ in the first 20 epochs and $10^{-3}$ after that. Cross entropy loss is used.
Highlights are generated by fans and consist of clips. We match each clip to when it happened in the full match and call this the “highlight clip” (non-overlapping).  The action of interest (“kill”, “objective control”, etc.) often happens in the later part of a highlight clip, while the clip contains some additional context before that action that may help set the stage.  For some of our experimental settings (Table~\ref{tab:ablation}), we used a heuristic of only including the last 25\% frames in every highlight clip as positive training examples. During evaluation, we used all frames in the highlight clip.

\paragraph{Ablation Study}
\label{sec:ablation}
Table~\ref{tab:ablation} shows the performance of each module separately on the dev set.  For the basic L-Char-LSTM and V-CNN models, using only the last 25\% of frames in highlight clips in training works best. In order to evaluate the performance of L-Char-LSTM model, we also train a Word-LSTM model by tokenizing all the chats and only considering the words that appeared more than 10 times, which results in 10019 words. We use this vocabulary to encode the words to 1-hot vectors. The L-Char-LSTM outperforms L-Word-LSTM by 22.3\%.

\begin{table}[t]
    \setlength{\tabcolsep}{6pt}
    \begin{tabular}{l|c|c|c}
    
    Method & Data & NALCS & LMS\\
    \hline
    L-Char-LSTM & chat& 43.2& 39.7 \\  
    \hline
    V-CNN-LSTM  & video& 72.2 & 69.2\\
    \hline
    $lv$-LSTM & \footnotesize{chat+video} & \textbf{74.7} & \textbf{70.0}\\
    \hline 
	\end{tabular}
    \caption{Test Results on the NALCS (English) and LMS (Traditional Chinese) datasets.} 
    \label{tab:final_results}
\end{table}

\paragraph{Test Results}
Test results are shown in Table~\ref{tab:final_results}. Somewhat surprisingly, the vision only model is more accurate than the language only model, despite the real-time nature of the comment stream. This is perhaps due to the visual form of the game,  where highlight events may have similar animations.  However, including language with vision in the $lv$-LSTM model significantly improves over vision alone, as the comments may exhibit additional contextual information.  Comparing results between ablation and the final test, it seems more data contributes to higher accuracy. This effect is more apparent in the vision models, perhaps due to complexity. Moreover, L-Char-LSTM performs better in English compared to traditional Chinese. From the numbers given in Section~\ref{sec:data_collection}, variation in the number of chats in NALCS was much higher than LMS, which one may expect to have a critical effect in the language model. However, our results seem to suggest that the L-Char-LSTM model can pickup other factors of the chat data (e.g. content) instead of just counting the number of chats. We expect a different language model more suitable for the traditional Chinese language should be able to improve the results for the LMS data.

\section{Conclusion}
We presented a new dataset and multimodal methods for highlight prediction, based on visual cues and textual audience chat reactions in multiple languages. We hope our new dataset can encourage further multilingual, multimodal research. 

\section*{Acknowledgments} 
We thank Tamara Berg, Phil Ammirato, and the reviewers for their helpful suggestions, and we acknowledge support from NSF 1533771.

\bibliography{emnlp2017}

\begin{thebibliography}{33}
\expandafter\ifx\csname natexlab\endcsname\relax\def\natexlab#1{#1}\fi

\bibitem[{Antol et~al.(2015)Antol, Agrawal, Lu, Mitchell, Batra, Zitnick, and
  Parikh}]{VQA}
Stanislaw Antol, Aishwarya Agrawal, Jiasen Lu, Margaret Mitchell, Dhruv Batra,
  C.~Lawrence Zitnick, and Devi Parikh. 2015.
\newblock {VQA}: {V}isual {Q}uestion {A}nswering.
\newblock In \emph{ICCV}.

\bibitem[{de~Avila et~al.(2011)de~Avila, Lopes, da~Luz~Jr., and
  de~A.~Araújo}]{VSUMM}
Sandra E.~F. de~Avila, Ana P.~B. Lopes, Antonio da~Luz~Jr., and Arnaldo
  de~A.~Araújo. 2011.
\newblock Vsumm: A mechanism designed to produce static video summaries and a
  novel evaluation method.
\newblock In \emph{Pattern Recognition Letters}.

\bibitem[{Bertini et~al.(2005)Bertini, Bimbo, and Nunziati}]{highlight_soccer}
M.~Bertini, A.~Del Bimbo, and W.~Nunziati. 2005.
\newblock Soccer videos highlight prediction and annotation in real time.
\newblock \emph{ICIAP}.

\bibitem[{Chen and Dolan(2011)}]{chen:acl11}
David~L. Chen and William~B. Dolan. 2011.
\newblock Collecting highly parallel data for paraphrase evaluation.
\newblock In \emph{ACL}.

\bibitem[{Cheng and Hsu(2006)}]{highlight_audio_basketball}
Chih{-}Chieh Cheng and Chiou{-}Ting Hsu. 2006.
\newblock Fusion of audio and motion information on hmm-based highlight
  extraction for baseball games.
\newblock \emph{{IEEE} Trans. Multimedia}.

\bibitem[{Chopra et~al.(2016)Chopra, Auli, and Rush}]{NLP_summary_abs}
Sumit Chopra, Michael Auli, and Alexander~M. Rush. 2016.
\newblock Abstractive sentence summarization with attentive recurrent neural
  networks.
\newblock In \emph{NAACL}.

\bibitem[{Filippova et~al.(2015)Filippova, Alfonseca, Colmenares, Kaiser, and
  Vinyals}]{NLP_summary_est2}
Katja Filippova, Enrique Alfonseca, Carlos~A Colmenares, Lukasz Kaiser, and
  Oriol Vinyals. 2015.
\newblock Sentence compression by deletion with lstms.
\newblock \emph{EMNLP}.

\bibitem[{Filippova and Altun(2013)}]{NLP_summary_est1}
Katja Filippova and Yasemin Altun. 2013.
\newblock The lack of parallel data in sentence compression.
\newblock \emph{EMNLP}.

\bibitem[{Fukui et~al.(2016)Fukui, Park, Yang, Rohrbach, Darrell, and
  Rohrbach}]{VQABilinear}
Akira Fukui, Dong~Huk Park, Daylen Yang, Anna Rohrbach, Trevor Darrell, and
  Marcus Rohrbach. 2016.
\newblock Multimodal compact bilinear pooling for visual question answering and
  visual grounding.
\newblock In \emph{EMNLP}.

\bibitem[{Graves(2013)}]{Char-LSTM}
Alex Graves. 2013.
\newblock Generating sequences with recurrent neural networks.
\newblock \emph{Neural computation}.

\bibitem[{Gygli et~al.(2014)Gygli, Grabner, Riemenschneider, and
  Van~Gool}]{SumMe}
Michael Gygli, Helmut Grabner, Hayko Riemenschneider, and Luc Van~Gool. 2014.
\newblock Creating summaries from user videos.
\newblock In \emph{ECCV}.

\bibitem[{He et~al.(2016)He, Zhang, Ren, and Sun}]{ResidualNetwork}
Kaiming He, Xiangyu Zhang, Shaoqing Ren, and Jian Sun. 2016.
\newblock Deep residual learning for image recognition.
\newblock In \emph{CVPR}.

\bibitem[{Hsieh et~al.(2012)Hsieh, Lee, Chiu, and Hsu}]{highlight_twitter}
Liang-Chi Hsieh, Ching-Wei Lee, Tzu-Hsuan Chiu, and Winston Hsu. 2012.
\newblock Live semantic sport highlight detection based on analyzing tweets of
  twitter.
\newblock \emph{ICME}.

\bibitem[{Huang et~al.(2016)Huang, Ferraro, Mostafazadeh, Misra, Devlin,
  Agrawal, Girshick, He, Kohli, Batra et~al.}]{visualstory}
Ting-Hao~K. Huang, Francis Ferraro, Nasrin Mostafazadeh, Ishan Misra, Jacob
  Devlin, Aishwarya Agrawal, Ross Girshick, Xiaodong He, Pushmeet Kohli, Dhruv
  Batra, et~al. 2016.
\newblock Visual storytelling.
\newblock In \emph{NAACL}.

\bibitem[{Kazemzadeh et~al.(2014)Kazemzadeh, Ordonez, Matten, and
  Berg}]{ReferItGame}
Sahar Kazemzadeh, Vicente Ordonez, Mark Matten, and Tamara Berg. 2014.
\newblock Referitgame: Referring to objects in photographs of natural scenes.
\newblock In \emph{EMNLP}.

\bibitem[{Krishna et~al.(2016)Krishna, Zhu, Groth, Johnson, Hata, Kravitz,
  Chen, Kalantidis, Li, Shamma, Bernstein, and Fei-Fei}]{visualgenome}
Ranjay Krishna, Yuke Zhu, Oliver Groth, Justin Johnson, Kenji Hata, Joshua
  Kravitz, Stephanie Chen, Yannis Kalantidis, Li-Jia Li, David~A Shamma,
  Michael Bernstein, and Li~Fei-Fei. 2016.
\newblock Visual genome: Connecting language and vision using crowdsourced
  dense image annotations.
\newblock In \emph{arXiv:1602.07332}.

\bibitem[{Lin et~al.(2014)Lin, Maire, Belongie, Hays, Perona, Ramanan, Dollár,
  and Zitnick}]{MSCOCO}
Tsung-Yi Lin, Michael Maire, Serge Belongie, James Hays, Pietro Perona, Deva
  Ramanan, Piotr Dollár, and C.~Lawrence Zitnick. 2014.
\newblock Microsoft coco: Common objects in context.
\newblock In \emph{ECCV}.

\bibitem[{Lu et~al.(2016)Lu, Yang, Batra, and Parikh.}]{attention_VQA}
Jiasen Lu, Jianwei Yang, Dhruv Batra, and Devi Parikh. 2016.
\newblock Hierarchical question-image co-attention for visual question
  answering.
\newblock In \emph{NIPS}.

\bibitem[{Mei et~al.(2016)Mei, Bansal, and Walter}]{NLP_summary_selective}
Hongyuan Mei, Mohit Bansal, and Matthew~R. Walter. 2016.
\newblock What to talk about and how? selective generation using lstms with
  coarse-to-fine alignment.
\newblock \emph{NAACL}.

\bibitem[{Nallapati et~al.(2016)Nallapati, Zhou, Gulcehre, Xiang
  et~al.}]{nallapati2016abstractive}
Ramesh Nallapati, Bowen Zhou, Caglar Gulcehre, Bing Xiang, et~al. 2016.
\newblock Abstractive text summarization using sequence-to-sequence rnns and
  beyond.
\newblock In \emph{CoNLL}.

\bibitem[{Ordonez et~al.(2011)Ordonez, Kulkarni, and Berg}]{im2text}
Vicente Ordonez, Girish Kulkarni, and Tamara~L Berg. 2011.
\newblock Im2text: Describing images using 1 million captioned photographs.
\newblock In \emph{NIPS}.

\bibitem[{Rashtchian et~al.(2010)Rashtchian, Young, Hodosh, and
  Hockenmaier}]{pascal_caption}
Cyrus Rashtchian, Peter Young, Micah Hodosh, and Julia Hockenmaier. 2010.
\newblock Collecting image annotations using amazon's mechanical turk.
\newblock \emph{NAACL HLT workshop}.

\bibitem[{Rohrbach et~al.(2015)Rohrbach, Rohrbach, Tandon, and
  Schiele}]{MovieDescription}
Anna Rohrbach, Marcus Rohrbach, Niket Tandon, and Bernt Schiele. 2015.
\newblock A dataset for movie description.
\newblock In \emph{CVPR}.

\bibitem[{Russakovsky et~al.(2015)Russakovsky, Deng, Su, Krause, Satheesh, Ma,
  Huang, Karpathy, Khosla, Bernstein, Berg, and Fei-Fei}]{ILSVRC15}
Olga Russakovsky, Jia Deng, Hao Su, Jonathan Krause, Sanjeev Satheesh, Sean Ma,
  Zhiheng Huang, Andrej Karpathy, Aditya Khosla, Michael Bernstein,
  Alexander~C. Berg, and Li~Fei-Fei. 2015.
\newblock Imagenet large scale visual recognition challenge.
\newblock In \emph{IJCV}.

\bibitem[{See et~al.(2017)See, Liu, and Manning}]{see2017get}
Abigail See, Peter~J Liu, and Christopher~D Manning. 2017.
\newblock Get to the point: Summarization with pointer-generator networks.
\newblock In \emph{ACL}.

\bibitem[{Song(2016)}]{yahoo_esports}
Yale Song. 2016.
\newblock Real-time video highlights for yahoo esports.
\newblock In \emph{arXiv:1611.08780}.

\bibitem[{Song et~al.(2015)Song, Vallmitjana, Stent, and Jaimes}]{TVSum}
Yale Song, Jordi Vallmitjana, Amanda Stent, and Alejandro Jaimes. 2015.
\newblock Tvsum: Summarizing web videos using titles.
\newblock In \emph{CVPR}.

\bibitem[{Sutskever et~al.(2014)Sutskever, Vinyals, and Le}]{wordLSTM}
Ilya Sutskever, Oriol Vinyals, and Quoc~V. Le. 2014.
\newblock Sequence to sequence learning with neural networks.
\newblock In \emph{NIPS}.

\bibitem[{Tapaswi et~al.(2016)Tapaswi, Zhu, Stiefelhagen, Torralba, Urtasun,
  and Fidler}]{MovieQA}
Makarand Tapaswi, Yukun Zhu, Rainer Stiefelhagen, Antonio Torralba, Raquel
  Urtasun, and Sanja Fidler. 2016.
\newblock Movieqa: Understanding stories in movies through question-answering.
\newblock In \emph{CVPR}.

\bibitem[{Torabi et~al.(2015)Torabi, Pal, Larochelle, and
  Courville.}]{MovieDescription2}
Atousa Torabi, Christopher Pal, Hugo Larochelle, and Aaron Courville. 2015.
\newblock Using descriptive video services to create a large data source for
  video annotation research.
\newblock In \emph{arXiv:1503.01070v1}.

\bibitem[{Wang et~al.(2004)Wang, Xu, Chng, and Tian}]{highlight_audio_soccer}
Jinjun Wang, Changsheng Xu, Engsiong Chng, and Qi~Tian. 2004.
\newblock Sports highlight detection from keyword sequences using hmm.
\newblock \emph{ICME}.

\bibitem[{Xiong et~al.(2016)Xiong, Merity, and Socher}]{memorynetwork_VQA}
Caiming Xiong, Stephen Merity, and Richard Socher. 2016.
\newblock Dynamic memory networks for visual and textual question answering.
\newblock In \emph{ICML}.

\bibitem[{Yu et~al.(2015)Yu, Park, Berg, and Berg}]{VisualMadlibs}
Licheng Yu, Eunbyung Park, Alexander~C. Berg, and Tamara~L. Berg. 2015.
\newblock Visual madlibs: Fill-in-the-blank image description and question
  answering.
\newblock In \emph{ICCV}.

\end{thebibliography}
\bibliographystyle{emnlp_natbib}

\end{document}